\documentclass[runningheads]{llncs}

\usepackage{eccv}

\usepackage{bm}
\usepackage{bbm}
\usepackage{amsmath}
\usepackage{amssymb}
\usepackage{eccvabbrv}

\usepackage{graphicx}
\usepackage{booktabs}

\usepackage[accsupp]{axessibility}  % Improves PDF readability for those with disabilities.

\usepackage{hyperref}

\usepackage{orcidlink}

\begin{document}

\title{Reducing Catastrophic Forgetting in Online Class Incremental Learning Using Self-Distillation} 

\titlerunning{OCIL Using Self-Distillation}

%TODO FINAL: Replace with your author list. 
\author{Kotaro Nagata\inst{1,2}\orcidlink{0000-0001-6732-0373} \and
Hiromu Ono\inst{1,3}\orcidlink{0009-0005-5487-3784} \and
Kazuhiro Hotta\inst{1,4}\orcidlink{0000-0002-5675-8713}}

%TODO FINAL: Replace with an abbreviated list of authors.
\authorrunning{K.~Nagata et al.}

\institute{Meijo University, Japan \and
%\email{\{180442097, 190442040\}@ccalumni.meijo-u.ac.jp} \and
\email{180442097@ccalumni.meijo-u.ac.jp} \and
\email{190442040@ccalumni.meijo-u.ac.jp} \and
\email{kazuhotta@meijo-u.ac.jp}}

\maketitle

\begin{abstract}
In continual learning, there is a serious problem of “ catastrophic forgetting”, in which previous knowledge is forgotten when a model learns new tasks. Various methods have been proposed to solve this problem. Replay methods which replay data from previous tasks in later training, have shown good accuracy. However, replay methods have a generalizability problem from a limited memory buffer. In this paper, we tried to solve this problem by acquiring transferable knowledge through self-distillation using highly generalizable output in shallow layer as a teacher.  
Furthermore, when we deal with a large number of classes or challenging data, there is a risk of learning not converging and not experiencing overfitting. Therefore, we attempted to achieve more efficient and thorough learning by prioritizing the storage of easily misclassified samples through a new method of memory update.
We confirmed that our proposed method outperformed conventional methods by experiments on CIFAR10, CIFAR100, and MiniimageNet datasets.
\end{abstract}

\section{Introduction}
\label{sec:intro}

In recent years, smart devices and image-related applications have been constantly generating a large amounts of image data. As data increases, AI models need to continually update the performance or be able to treat many tasks. This kind of such a learning method is called continual learning\cite{de2021continual}. This enables the learning of an intelligence like mammals. Among them, more practical continual learning using streaming data called online continual learning\cite{gepperth2016incremental, losing2018incremental}. In this paper, we handle a online class incremental continual learning, which is a setup of gradually increasing the number of classes.

There is a serious problem of forgetting old knowledge when AI model tries to learn a new task, called ”catastrophic forgetting”\cite{mccloskey1989catastrophic, goodfellow2013empirical}. To mitigate catastrophic forgetting, there are various methods to store the previous task information. Replay methods\cite{rolnick2019experience, chaudhry2018efficient, aljundi2019gradient, NEURIPS2019_15825aee, shim2021online, prabhu2020gdumb, mai2021supervised, rebuffi2017icarl} store a small portion of past samples and replay the samples along with present task samples. Regularization-based methods\cite{li2017learning, rolnick2019experience} update CNN’s parameters based on how important it is to previous tasks. Parameter isolation methods\cite{mallya2018packnet, rusu2016progressive} expand the networks or decompose the network into subnetworks for each task. Among the recently proposed approaches, replay methods is one of the most effective methods for mitigating catastrophic forgetting\cite{mai2022online}.

However, replay methods suffer from a problem where the limited memory buffer results in fewer learning samples from past tasks, leading to overfitting. Considering real-world environments, it is generally preferred for the memory buffer of Replay methods to be small, making this problem critical. Replay methods store samples from tasks during learning in the memory buffer, but they face the problem of repeatedly learning easily identifiable samples. Therefore, in this paper, we improved the generalization capability of our model by incorporating a self-distillation mechanism\cite{hinton2015distilling}, where the shallow layers of the neural network contain general knowledge and the deeper layers contain specialized knowledge. By distilling the features of the shallow layers into the deeper layers of the same model, we enhanced its generalization performance. Furthermore, with the new memory update method, we prioritize saving $n$ images with a low probability of the correct class by the classifier, thereby prioritizing classes prone to errors. This enables more efficient learning and ensures thorough training.

In experiments, we used CIFAR10, CIFAR100\cite{krizhevsky2009learning} and MiniImageNet\cite{vinyals2016matching} to validate our method. As a result, the proposed method showed a reduction in catastrophic forgetting compared to several conventional online continual learning methods. Especially, for the smallest buffer sizes (M=100,500,500) on CIFAR 10, CIFAR100 and MiniImageNet, the maximum improvement in classification accuracy for each was $5.9\%$, $3.2\%$, and $4.0\%$ in comparison with baseline method.

This paper is organized as follows. We describe related works in section 2. Our proposed method is explained in section 3. Section 4 is for experimental results. Finally, conclusions and future works are described in section 5.

\section{Related works}

\subsection{Continual learning scenario}

There are many continual learning setups that a neural network model needs to sequentially learn a series of tasks. In this paper, we categorize them into three setups, task-incremental(Task-IL), class-incremental(ClassIL) and domain-incremental learning(Domain-IL), depending on whether the task-ID is given at the test time\cite{van2019three}. Task-IL are always informed about which task needs to be performed, called multi-head setup. This is the easiest continual learning scenario. Domain-IL cannot use task-ID at the test time. Models only need to solve the task at hand; they are not required to infer which task it is. In contrast to task IL, in class IL, the model is not given a taskID and must be able to both solve each task we have seen and guess which task it is. The class-IL is more challenging than task-IL and domain-IL, but also more realistic. Therefore, in this paper, we conduct experiments in the more practical setup of Class-IL.

\subsection{Replay methods in continual learning}

Continual learning methods are mainly classified into three mechanisms for mitigating catastrophic forgetting: repay methods\cite{rolnick2019experience, chaudhry2018efficient, aljundi2019gradient, NEURIPS2019_15825aee, shim2021online, prabhu2020gdumb, mai2021supervised, rebuffi2017icarl}, regularization-based methods\cite{li2017learning, rolnick2019experience} and parameter isolation methods \cite{mallya2018packnet, rusu2016progressive}. 
Replay methods store a portion of previous tasks samples and update to replay past samples. Regularization-based methods restrict the parameters of the model 
so that it does not move away from the parameters of past tasks. 
Parameter isolation methods reduce forgetting by assigning model parameters to each task or by extending the model. Among them, replay methods has shown great performance in continual learning, despite the simple methods. In replay methods, Experience Replay (ER) is a simple framework with buffering past samples and a tuned learning rate scheduling to prevent forgetting past knowledge. Many methods have been proposed based on ER in terms of how to store samples and how to use them. Among these, contrastive learning\cite{khosla2020supervised, chen2020simple} is also considered effective for continual learning. However, in replay methods, the capacity of the buffer is limited, resulting in a small amount of data from past tasks that can be stored. Consequently, the effectiveness of contrastive learning may not be fully realized due to insufficient use of past images and there is a problem of overfitting, as the model may fail to acquire highly generalized knowledge by repeatedly learning images in the buffer during subsequent tasks. When we deal with a large number of classes or challenging data, the learning process may not converge, leading to a possibility of not experiencing overfitting. Therefore, in this paper, we leverage the nature of neural networks and incorporate a self-distillation mechanism to improve generalization. Furthermore, with the new memory update method, we aim to achieve more efficient learning and ensure thorough training.

\section{Proposed Method}

\subsection{Motivation}

Among the conventional continual learning algorithms, replay methods have shown great performance\cite{rolnick2019experience, chaudhry2018efficient, aljundi2019gradient, NEURIPS2019_15825aee, shim2021online, prabhu2020gdumb, mai2021supervised, rebuffi2017icarl}. Replay methods store a portion of the past training samples in a memory buffer of fixed capacity and replay them in a later task. However, because the buffer capacity is fixed, the varieties of samples for each past task decreased as the task progresses. Consequently, there is a problem of overfitting, as the model may fail to acquire highly generalized knowledge by repeatedly learning images in the buffer during subsequent tasks. In continual learning, the effectiveness of contrastive learning may not be fully realized due to insufficient use of past images. 

To solve these problems, we proposed two elements. 
\def\labelenumi{(\theenumi)}
\begin{enumerate}
    \item We incorporate a self-distillation loss$\mathcal{L}_{dist}^{self}$ to improve generalization.  
    The self-distillation loss enables us to leverage highly generalized features in the shallow layers of the AI model.
    \item We introduce the new memory update method that prioritizes the storage of easily mistaken samples. The new memory update method can enable more efficient learning and ensure sufficient learning.
\end{enumerate}

\subsection{Problem setting}

In this section, we conduct problem setting for online Class-IL. In online Class-IL, tasks are continuously learned in the data stream $D = \{D_1, D_2, ..., D_T\}$. Here, $D_t = \{x_i, y_i\}_{i=1}^{N_t}$ represents the dataset for task $t$, where $T$ represents the total number of tasks. Dataset $D_t$ represents the number of labeled samples $N_t$, and $y_i$ represents the class label of sample $x_i$. Here, $y_i$ is expressed as $y_i \in C_t$ using the set of classes $C_t$ included in task $t$. In replay methods, a portion of the learned samples is stored in a fixed-capacity memory buffer $M$ and at each time step of task $t$, $X \cup X^b$ is input to the model. Here, $X$ and $X_b$ represent samples taken from the data distribution $D$ and memory buffer $M$, respectively. The goal of online Class-IL is to achieve higher classification accuracy for all classes when all data have been learned once.

\begin{figure*}
  \centering
  \includegraphics[height=42mm, width=0.60\linewidth]{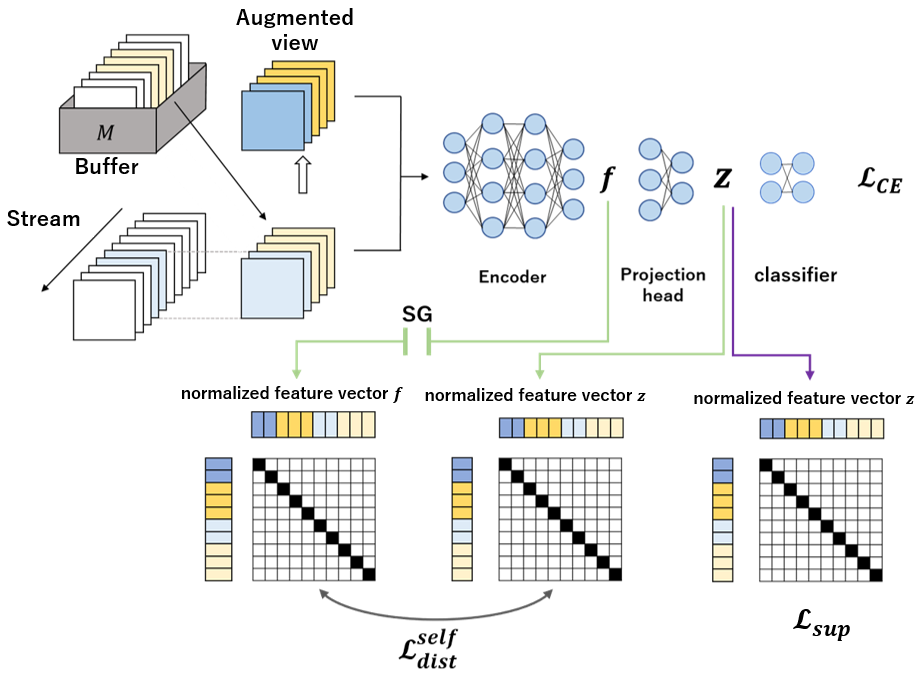}
  \caption{The overview of the proposed method. Our approach is based on Supervised Contrastive Replay (SCR) of Replay method. Distilling knowledge from shallow layers by aligning the similarity maps of normalized features.}
  \label{fig:short-a}
\end{figure*}

\subsection{Overview of the proposed method}

The overview of the proposed method is shown in Figure 1. Our approach is inspired by the method called Supervised Contrastive Replay (SCR)\cite{mai2021supervised}, which is a form of Replay method. It consists of an Encoder, a Projection head, and a Classifier structure to facilitate identification. The output of the Encoder is denoted as $f$, the output of the Projection head as $z$, and the learning is conducted using the following loss function $\mathcal{L}$.
\begin{equation}
  \mathcal{L} = \mathcal{L}_{sup} + \mathcal{L}_{ce} + \mathcal{L}_{dist}^{self}
  \label{eq:important}
\end{equation}
where $\mathcal{L}_{sup}$ conducts contrastive learning using the normalized embedding vector $z$ to learn the relationship between the image features of different classes.

$\mathcal{L}_{sup}$ is represented by the following equation.
\begin{equation}
  \mathcal{L}_{sup} = \sum_{i=1}^{2N} \frac{-1}{|P(i)|} \sum_{p\in P(i)} 
  \log{\frac{exp(z_i\cdot{z_p}/{\tau})}{\sum_{j\in A(i)}exp(z_i\cdot{z_a}/{\tau})}}
  \label{eq:important}
\end{equation}
where $i$ is an anchor and the input mini-batch consists of a total of 2$N$ images, including original samples and augmented samples. The original samples consist of samples $X$ extracted from the stream and samples $X_b$ extracted from the buffer. Additionally, $P(i)$ represents positive examples with the same label as the anchor, while $A(i)$ represents images different from the anchor.

\subsection{Self-distillation mechanism}

Replay methods cause overfitting by training on a small number of images stored in the memory buffer. To alleviate the overfitting, this paper proposes a self-distillation loss, which distills the highly generalized knowledge from the shallow layers of the network into the deeper layers \cite{szegedy2016rethinking}.
In self-distillation, knowledge distillation is performed by bringing the relationship between the features of the deep layers closer to the relationship between the features of the shallow layers within the samples of the mini-batch because the dimension of features at shallow and deep layers is different.
The relationship between the features is represented as follows.
%\begin{equation}
  ${p(z_i)} = [p_{i, 1}, ..., p_{i, 2N}]$
%  \label{eq:important}
%\end{equation}
where $p_{i, j}$ represents cosine similarity between normalized feature vectors as
\begin{equation}
  p_{i, j} = \frac{exp(z_i\cdot{z_p}/{\kappa})}{\sum_{k \neq i}exp(z_i\cdot{z_k}/{\kappa})}
  \label{eq:important}
\end{equation}
where $i$ is excluded because the cosine similarity with itself is always equal to 1.
where $\kappa$ is a temperature hyperparameter. Using the cosine similarity vector of normalized feature vectors, $\mathcal{L}_{dist}^{self}$ is expressed as
\begin{equation}
  \mathcal{L}_{dist}^{self} = \sum_{i=1}^{2N}-p(z_i)\cdot log p(f_i)
  \label{eq:important}
\end{equation}
where $f$ is the features of the projection head.

We perform knowledge distillation by minimizing the KL divergence between the cosine similarity vectors of feature vectors outputted from the shallow layers of the Encoder in the mini-batch and those outputted from the deep layers of the Projection head.

\begin{figure*}
  \centering
  \includegraphics[height=30mm, width=0.8\linewidth]{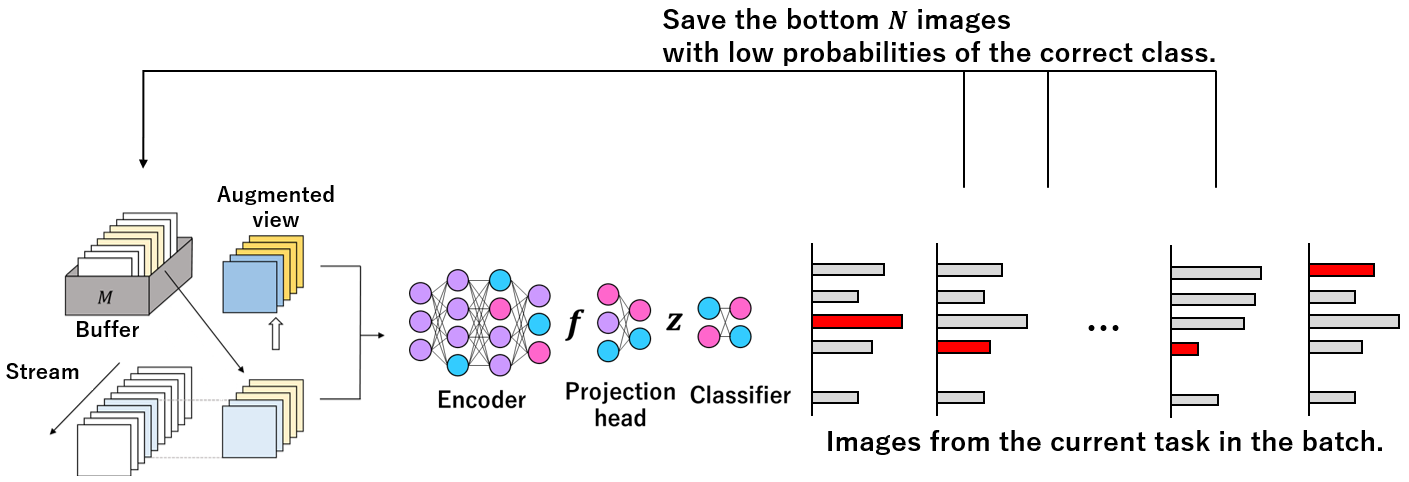}
  \caption{The new memory update method. Save the bottom $N$ images with low probabilities of the correct class (in this experiment, $N$=5) and prioritize storing them.}
  \label{fig:short-a}
\end{figure*}

\subsection{Memory update method}

Self-distillation addresses the issue of overfitting in Replay methods by improving generalization. However, when we deal with a large number of classes or challenging data, the learning may not converge, and there is a risk of overfitting. Therefore, in this paper, we propose a new memory update method that prioritizes the storage of easily misclassified samples to achieve more efficient and thorough learning. Conventionally, samples from the current task were stored in the memory buffer, which could lead to the repetition of learning easy-to-discriminate samples. Therefore, we prioritize storing samples from easily misclassified classes by storing $N$ ($N$=5 in experiment) images with low probabilities of the correct class by the classifier in Figure 2.

\subsection{Inference method}

During inference, similar to SCR, we compute the average of the features $f$ stored in the buffer and use the Nearest Class Mean Classifier (NCM classifier)\cite{mensink2013distance, mai2021supervised} to predict the label of the test image based on the nearest prototype's label. The equation for the NCM classifier can be represented as 
\begin{equation}
  \mu_c = \frac{1}{n_c}\sum_{i}Enc(x_i)\cdot \mathbbm{1} \{y_i = c\}
  \label{eq:important}
\end{equation}
\begin{equation}
  y^* = \underset{c\in C_t} {\operatorname{argmin}} \| {Enc(x)-\mu_c} \|
  \label{eq:important}
\end{equation}
where $n_c$ is the number of samples in the memory buffer for class $c$ and {$y_i = c$} is the indicator for $y_i = c$. The prototype $\mu_c$ is the centroid of the embedding of the samples of each class in the buffer. The prototype is recomputed at each inference step using the samples in the buffer at that time.

\section{Experiments}

\subsection{Experiment Setup}

%\subsubsection{Datasets}

{\bf Datasets:}
We conducted experiments on three datasets: Split CIFAR10/100\cite{krizhevsky2009learning} and Split MiniimageNet\cite{vinyals2016matching}. Split CIFAR10 divides CIFAR10 into 5 tasks, each task consists of disjoint 2 classes. Split CIFAR100 and Split MiniimageNet split them into 10 tasks, each task consists of disjoint 10 classes.\\
%\subsubsection{Comparison methods}
{\bf Comparison methods:}
To validate the effectiveness of our method, we compare our method with several continual learning methods: ER\cite{rolnick2019experience}, EWC\cite{rolnick2019experience}, LwF\cite{li2017learning}, ASER\cite{shim2021online}, AGEM\cite{chaudhry2018efficient}, MIR\cite{NEURIPS2019_15825aee}, GSS\cite{aljundi2019gradient}, GDumb\cite{prabhu2020gdumb}, iCaRL\cite{rebuffi2017icarl}, SCR\cite{mai2021supervised}. We also evaluated offline and fine-tuning. 
Offline is not a continual learning setting, but trains model in multiple epochs on the whole dataset with iid sampled mini-batches. Fine-tuning trains models in a continual learning setting without measures against catastrophic forgetting.\\
%\subsubsection{Evaluation Metric}
{\bf Evaluation metric:}
In this experiments, we used Average Accuracy $A_i$ as the evaluation metric\cite{lesort2020continual}. $A_i$ can be represented as
%\begin{equation}
$A_{i, j} = \frac{1}{i}\sum_{j=1}^ia_{i, j}$
% \label{eq:important}
%\end{equation}
where $a_{i, j}$ represents the accuracy on task $j$ after learning task $i$. In this paper, we use the average accuracy $A_T$ of all tasks at the end of all tasks.\\
%\subsubsection{Experimental Details}
{\bf Experimental details:}
In experiments on all datasets, we used ResNet18\cite{he2016deep} as the backbone, SGD as the optimizer, $0.01$ as the learning rate, and $1.0 \times 10^{-4}$ as the decay rate. In the Replay Methods, 10 samples are randomly retrieved from the data stream and 100 samples are randomly retrieved from the buffer to form 110 mini-batches. For SCR and the proposed method, the feature vector of 128 dimensions was output by MLP using the activation function ReLU as the projection head, and NCM was used for classification. For offline, we adopted 50 epochs as training. We use reservoir sampling\cite{vitter1985random} for memory update and random sampling for memory retrieval. Additionally, the experimental results were obtained by conducting experiments with the order of classes to be learned randomly changed 10 times, and using the average accuracy of those results.
Furthermore, we conducted experiments with the order of learning classes randomly changed 10 times, and the average accuracy of these experiments was used.

\begin{table*}[h]
 \caption{Comparison results on Split CIFAR10, CIFAR100 and MiniImageNet. All scores are Average Accuracy by the end of training and average of 10 runs. $M$ is a buffer size. The best scores are in boldface and the second best scores are underlined.}
 \label{table:ex_booktabs}
 \centering
  \scalebox{0.55}[0.78]{
  \begin{tabular}{lcccccccccccc}
   \toprule              % --- booktabs: 上の罫線 ---
   Method & \multicolumn{4}{c}{CIFAR10} & \multicolumn{4}{c}{CIFAR100} & \multicolumn{4}{c}{MiniImageNet} \\
   \cmidrule(lr){2-5}    % --- booktabs: 2 から 5 列の罫線 ---
   \cmidrule(lr){6-9}
   \cmidrule(lr){10-13}
       & M=100 & M=200 & M=500 & M=1000 & M=500 & M=1000 & M=2000 & M=5000 & M=500 & M=1000 & M=2000 & M=5000 \\
   \hline\hline              % --- booktabs: 中間の二重罫線 ---
   offline & \multicolumn{4}{c}{$81.7\pm0.5$} & \multicolumn{4}{c}{$50.1\pm0.3$} & \multicolumn{4}{c}{$51.6\pm0.4$} \\
   finetuning & \multicolumn{4}{c}{$17.5\pm1.1$} & \multicolumn{4}{c}{$4.7\pm0.5$} & \multicolumn{4}{c}{$4.5\pm0.5$} \\
   EWC\cite{rolnick2019experience} & \multicolumn{4}{c}{$17.5\pm1.3$} & \multicolumn{4}{c}{$4.7\pm0.6$} & \multicolumn{4}{c}{$4.6\pm0.7$}\\
   LwF\cite{li2017learning} & \multicolumn{4}{c}{$22.3\pm0.8$} & \multicolumn{4}{c}{$12.9\pm0.5$} & \multicolumn{4}{c}{$11.2\pm0.9$} \\
   \midrule              % --- booktabs: 中間の罫線 ---
   ER\cite{rolnick2019experience} & $20.8\pm1.2$ & $21.6\pm1.8$ & $28.3\pm3.5$ & $36.1\pm4.3$ & $9.3\pm1.2$ & $12.2\pm1.1$ & $15.5\pm1.4$ & $20.6\pm1.8$ & $8.4\pm0.9$ & $10.9\pm0.7$ & $14.4\pm0.9$ & $17.7\pm2.3$ \\
   ASER\cite{shim2021online} & $19.3\pm0.9$ & $21.4\pm1.6$ & $26.1\pm3.0$ & $31.9\pm3.3$ & $11.7\pm1.3$ & $14.7\pm1.0$ & $18.8\pm0.7$ & $23.9\pm1.3$ & $10.8\pm0.9$ & $12.6\pm1.1$ & $14.0\pm1.3$ & $18.8\pm4.3$ \\
   A-GEM\cite{chaudhry2018efficient} & $18.6\pm0.9$ & $17.8\pm1.5$ & $18.1\pm1.1$ & $18.1\pm1.3$ & $5.4\pm0.6$ & $5.4\pm0.6$ & $5.6\pm0.6$ & $5.7\pm0.6$ & $5.1\pm0.3$ & $4.9\pm0.4$ & $4.7\pm0.7$ & $5.0\pm0.7$ \\
   MIR\cite{NEURIPS2019_15825aee} & $20.4\pm0.6$ & $22.3\pm2.0$ & $29.2\pm2.4$ & $37.1\pm3.7$ & $9.3\pm0.8$ & $11.5\pm1.5$ & $15.7\pm1.0$ & $22.0\pm1.8$ & $8.3\pm0.5$ & $10.3\pm0.7$ & $14.9\pm0.8$ & $18.3\pm2.3$ \\
   GSS\cite{aljundi2019gradient} & $18.7\pm1.1$ & $20.1\pm0.8$ & $24.8\pm1.3$ & $31.5\pm4.0$ & $8.6\pm0.8$ & $9.8\pm0.7$ & $13.3\pm0.8$ & $16.0\pm1.5$ & $8.1\pm0.9$ & $9.9\pm0.6$ & $13.1\pm1.7$ & $15.1\pm1.9$ \\
   GDumb\cite{prabhu2020gdumb} & $22.9\pm1.4$ & $27.1\pm1.6$ & $32.4\pm1.4$ & $37.5\pm1.3$ & $7.0\pm0.5$ & $9.9\pm0.4$ & $13.3\pm0.6$ & $19.3\pm0.5$ & $5.3\pm0.5$ & $7.3\pm0.8$ & $11.8\pm0.6$ & $20.5\pm0.7$ \\
   iCaRL\cite{rebuffi2017icarl} & $26.8\pm2.8$ & $30.8\pm2.4$ & $38.2\pm3.1$ & $49.6\pm2.8$ & $13.3\pm0.9$ & $16.4\pm0.7$ & $18.6\pm0.6$ & $19.1\pm0.6$ & $10.4\pm0.8$ & $12.6\pm0.6$ & $14.2\pm0.7$ & $15.7\pm0.9$ \\
   SCR\cite{mai2021supervised} & \underline{$35.1\pm2.9$} & \underline{$45.4\pm1.7$} & \underline{$57.4\pm1.0$} & \underline{$64.5\pm1.2$} & \underline{$19.3\pm0.6$} & \underline{$26.4\pm0.5$} & \underline{$32.7\pm0.6$} & \underline{$38.6\pm0.5$} & \underline{$17.8\pm1.2$} & \underline{$24.3\pm0.7$} & \underline{$31.0\pm1.1$} & \underline{$35.8\pm0.8$} \\
   \midrule              % --- booktabs: 中間の罫線 ---
   ours & $\bm{41.0\pm3.1}$ & $\bm{49.5\pm2.4}$ & $\bm{58.5\pm1.0}$ & $\bm{64.9\pm0.8}$ & $\bm{22.5\pm0.6}$ & $\bm{28.4\pm0.5}$ & $\bm{33.9\pm0.8}$ & $\bm{39.8\pm0.7}$ & $\bm{21.8\pm0.5}$ & $\bm{26.6\pm0.5}$ & $\bm{31.5\pm1.0}$ & $\bm{36.5\pm0.6}$ \\
   \bottomrule           % --- booktabs: 底の罫線 ---
  \end{tabular}
  }
\end{table*}

\subsection{Comparison results}

We compare our method with various online continual learning methods on Split CIFAR10, Split CIFAR100 and Split MiniImageNet in Table 1.
We evaluate the accuracy at the end of training for multiple datasets at various buffer sizes. 
SCR exhibits the highest performance among the baselines across various buffer sizes. This is because contrastive learning and NCM classifiers are effective in biasing model weights from class imbalance between past and current classes. 
However, our method outperforms top-performing baseline SCR in accuracy across all buffer sizes.
Specifically, for the smallest buffer sizes (M=100,500,500) on all datasets, the accuracy increases by $5.9\%$, $3.2\%$, and $4.0\%$. This indicates a significant improvement when the number of stored samples is low, suggesting a mitigation of overfitting due to the limited number of stored samples.

\begin{table*}[h] 
 \caption{Ablation study of two components in our method: self-distillation and buffer processing. The best scores are in boldface and the second best scores are underlined.}
 \label{table:ex_booktabs}
 \centering
  \scalebox{0.52}[0.77]{
  \begin{tabular}{lcccccccccccc}
   \toprule              % --- booktabs: 上の罫線 ---
   Method & \multicolumn{4}{c}{CIFAR10} & \multicolumn{4}{c}{CIFAR100} & \multicolumn{4}{c}{MiniImageNet} \\
   \cmidrule(lr){2-5}    % --- booktabs: 2 から 5 列の罫線 ---
   \cmidrule(lr){6-9}
   \cmidrule(lr){10-13}
       & M=100 & M=200 & M=500 & M=1000 & M=500 & M=1000 & M=2000 & M=5000 & M=500 & M=1000 & M=2000 & M=5000 \\
   \hline\hline              % --- booktabs: 中間の二重罫線 ---
   $\mathcal{L}_{sup}$ & $37.7\pm1.0$ & $46.3\pm2.4$ & $57.6\pm1.5$ & $64.4\pm1.0$ & $18.7\pm0.7$ & $26.3\pm0.8$ & $32.7\pm0.6$ & $38.3\pm0.5$ & $17.4\pm1.2$ & $24.6\pm0.9$ & $30.2\pm0.9$ & $35.4\pm0.8$ \\
   $\mathcal{L}_{sup} + \mathcal{L}_{ce}$ & $38.1\pm2.6$ & $47.3\pm2.6$ & $58.4\pm1.4$ & $64.8\pm1.0$ & $19.7\pm0.9$ & $27.0\pm0.7$ & $33.8\pm0.8$ & \underline{$40.6\pm0.5$} & $18.8\pm0.7$ & $25.2\pm0.8$ & $31.3\pm0.6$ & $\bm{38.1\pm0.6}$ \\
   $\mathcal{L}_{sup} + \mathcal{L}_{ce} + \mathcal{L}_{dist}^{self}$ & \underline{$40.0\pm2.3$} & \underline{$48.8\pm2.0$} & $\bm{58.9\pm1.6}$ & $\bm{65.0\pm1.2}$ & $20.1\pm0.6$ & $27.1\pm0.7$ & $\bm{34.1\pm0.8}$ & $\bm{41.2\pm0.4}$ & $18.0\pm0.8$ & $24.1\pm1.0$ & $30.0\pm0.8$ & $37.6\pm1.0$ \\
   $\mathcal{L}_{sup} + \mathcal{L}_{ce} + \text{buffer}$ & $38.8\pm3.4$ & $46.9\pm2.3$ & $57.9\pm1.5$ & $64.0\pm0.8$ & \underline{$22.5\pm0.5$} & $\bm{28.5\pm0.5}$ & $33.9\pm0.7$ & $39.7\pm0.4$ & \underline{$21.7\pm0.8$} & $\bm{27.5\pm0.9}$ & $\bm{32.5\pm1.0}$ & \underline{$38.1\pm0.4$} \\
   \midrule              % --- booktabs: 中間の罫線 ---
   ours & $\bm{41.0\pm3.1}$ & $\bm{49.5\pm2.4}$ & \underline{$58.5\pm1.0$} & \underline{$64.9\pm0.8$} & $\bm{22.5\pm0.6}$ & \underline{$28.4\pm0.5$} & \underline{$33.9\pm0.8$} & $39.8\pm0.7$ & $\bm{21.8\pm0.5}$ & \underline{$26.6\pm0.5$} & \underline{$31.5\pm1.0$} & $36.5\pm0.6$ \\
   \bottomrule           % --- booktabs: 底の罫線 ---
  \end{tabular}
  }
\end{table*}

\subsection{Ablation Study}

This section shows the effectiveness of each element of the proposed method. Table 2 compares our method with various types of loss functions. From the results, both self-distillation and buffer processing are effective in most cases. Particularly, self-distillation is effective when the buffer size is small, while buffer processing is effective for Miniimagenet, which is difficult to converge during training. Moreover, it can be seen that the highest improvement in accuracy occurs when both methods are introduced. This suggests that buffer processing enables more efficient learning, while self-distillation enhances generalization, leading to complementary learning between the two methods.

\section{Conclusions}

We focused on the problem of having a limited 
number of past training images in continual learning based on Replay methods. To address the decrease 
in generalization caused by overfitting, we attempted to improve the generalization by conducting self-distillation using highly generalized features in shallow layer as teachers. 
We also proposed a new memory update method that prioritizes the storage of easily misclassified samples to achieve more efficient 
and thorough learning. As a result, we observed a maximum improvement of $5.9\%$ compared to the baseline method. Currently, knowledge distillation is performed only from a single layer, so we aim to explore distillation from multiple layers and investigate more effective methods for sample preservation.

\clearpage  % TODO REVIEW/FINAL: This \clearpage needs to be removed from both review and camera-ready versions.

\bibliographystyle{splncs04}
\bibliography{main}
\end{document}